\documentclass{article}

\usepackage[final]{nips_2016}

\usepackage[utf8]{inputenc} % allow utf-8 input
\usepackage[T1]{fontenc}    % use 8-bit T1 fonts
\usepackage{hyperref}       % hyperlinks
\usepackage{url}            % simple URL typesetting
\usepackage{booktabs}       % professional-quality tables
\usepackage{amsfonts}       % blackboard math symbols
\usepackage{nicefrac}       % compact symbols for 1/2, etc.
\usepackage{microtype}      % microtypography
\usepackage{setspace}
\usepackage{epsfig}
\usepackage{todonotes}
\usepackage{float}

\usepackage{subfigure}
\usepackage{graphicx}
\usepackage{caption}
\usepackage{booktabs}
\usepackage{tikz}
\usepackage{array}

\newcolumntype{M}[1]{>{\centering\arraybackslash}m{#1}}

\title{Incorporating Human Domain Knowledge \\into Large Scale Cost Function Learning}
\author{
Markus Wulfmeier \\
Oxford Robotics Institute\\
Department of Engineering Science\\
University of Oxford\\
\texttt{markus@robots.ox.ac.uk} \\
\And
Dushyant Rao \\
Oxford Robotics Institute\\
Department of Engineering Science\\
University of Oxford\\
\texttt{dushyant@robots.ox.ac.uk} \\
\AND
Ingmar Posner \\
Oxford Robotics Institute\\
Department of Engineering Science\\
University of Oxford\\
\texttt{ingmar@robots.ox.ac.uk} \\
}
\PassOptionsToPackage{square,sort,comma,numbers}{natbib}
\begin{document}

\maketitle

\begin{abstract}
Recent advances have shown the capability of Fully Convolutional Neural Networks (FCN) to model cost functions for motion planning in the context of learning driving preferences purely based on demonstration data from human drivers. While pure learning from demonstrations in the framework of Inverse Reinforcement Learning (IRL) is a promising approach, we can benefit from well informed human priors and incorporate them into the learning process.
%we can benefit from applying prior structure from well informed human priors in many areas.
%Our work targets refining cost functions for path planning by
Our work achieves this by pretraining a model to regress to a manual cost function and refining it based on Maximum Entropy Deep Inverse Reinforcement Learning. When injecting prior knowledge as pretraining for the network, we achieve higher robustness, more visually distinct obstacle boundaries, and the ability to capture instances of obstacles that elude models that purely learn from demonstration data. Furthermore, by exploiting these human priors, the resulting model can more accurately handle corner cases that are scarcely seen in the demonstration data, such as stairs, slopes, and underpasses.
\end{abstract}

\section{Introduction}
Manual handcrafting of cost functions for motion planning systems is an inherently complex and time consuming task. It requires high competency in the target area and expert knowledge about robotics systems and the applied algorithms. Ideally, robotic behaviour can be defined by untrained personnel, enabling task adaptation without involving highly trained experts. Inverse Reinforcement Learning (IRL) targets this problem by learning direct reward models from demonstration samples, and has been successfully applied to problems in a wide range of areas \cite{WulfmeierIROS2016,FinnLA16,NIPS2015_5882}.

Recent advances exploit the ease of generating samples for learning from demonstration, and combining with with high capacity representations through Neural Networks in domains such as games \cite{silver2016mastering} and autonomous driving \cite{WulfmeierIROS2016}.
%The following sentence is really clunky
% Combinations of learning from demonstration with high capacity models have been part of recent advances in domains such as games \cite{silver2016mastering} and autonomous driving \cite{WulfmeierIROS2016}. 
While the possibility to create large amounts of training data without manual labelling enabled training large networks, corner cases still represent a challenge for deep learning as less training data is present. Our work targets initialising neural networks by employing human priors to improve performance in these cases.

\begin{figure}
	\centering
	\includegraphics[width = 0.55\textwidth]{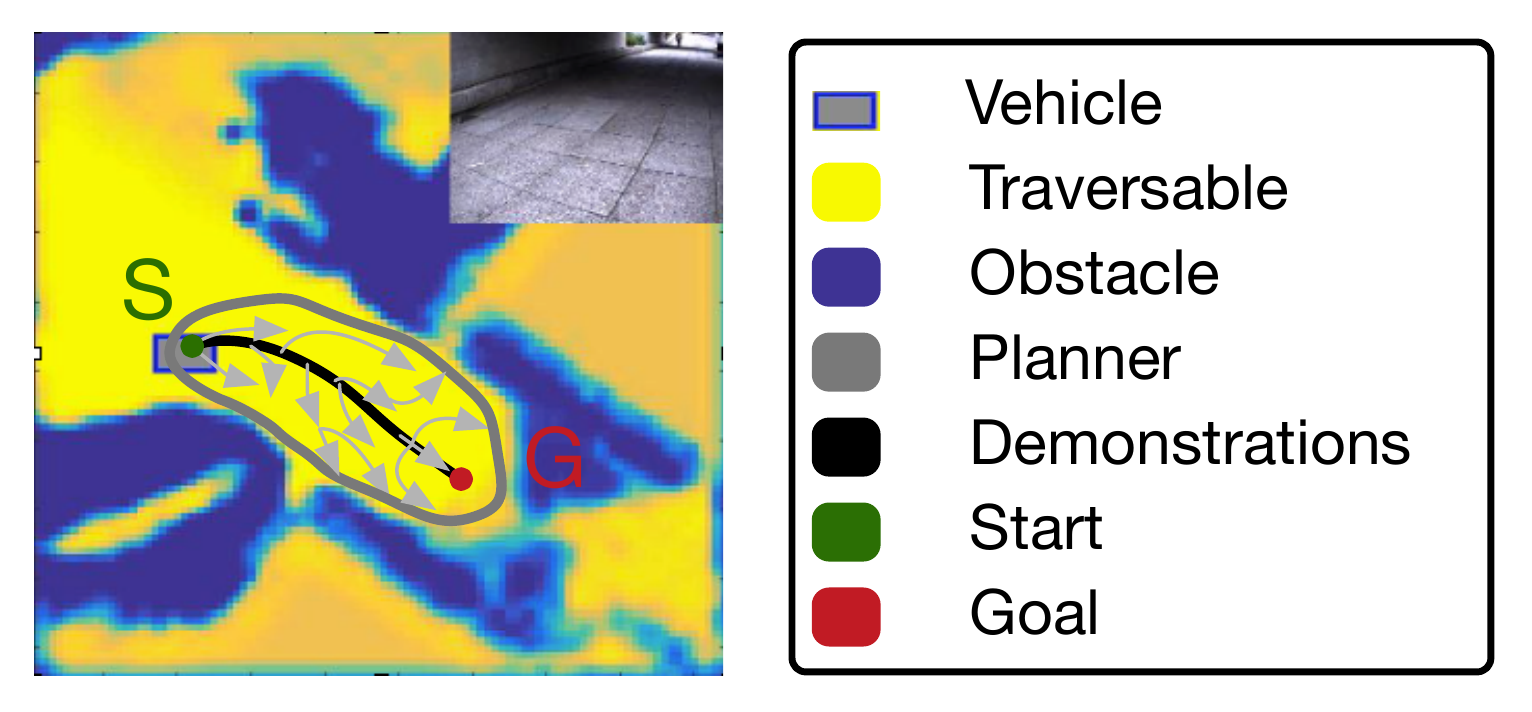}
	\caption{Illustration of sparse feedback, showing a demonstration trajectory on the spatial cost map around the vehicle, as well as the region explored by the planning algorithm. Error feedback is only created for the area surrounding sample trajectories.} 
	\label{fig:sparse}
\end{figure}

While CNNs can be integrated straightforwardly into different domains, the IRL framework introduces additional challenges. One of the dominant influences is spatially sparse feedback from our objective function as displayed in Figure \ref{fig:sparse}. For example, when training a model for image segmentation, the objective creates feedback for each individual pixel. When training with a loss based on Deep Inverse Reinforcement Learning on the other hand, error terms will focus on the region around the demonstration trajectories. These error terms are based on states visited by the demonstration samples and the planning algorithm, which inherently focuses around sample data. Our work addresses the shortcomings by pretraining the network towards a dense human-provided prior, to learn richer feature representations for untraversed areas and increase the network's ability to generalise. 

%One drawback of learning from demonstration is the fact that driving datasets contain very few, if any, examples of collisions and driving behind untraversable boundaries, which means that an IRL approach inherently will focus on features for traversed terrain and areas close to the visited states. 

%Thus, the ability to explicitly incorporate human priors into learned cost maps is an important one, as it affords the ability to explicitly encode which regions are untraversable and which behaviours the planner must avoid. markus: we just dont do that since we completely refine them :) 

% While pure \textbf{auto-encoder} pretraining will provide knowledge about features describing the complete encountered terrain, we follow a more promising approach and train the network as regression model towards an educated guess of the cost map given by the manual cost function currently incorporated into our planning system. Due to similarity between pretraining and IRL training, the representations extracted by this approach are of higher significance for the final task.
% Furthermore, the approach directly applies the same network architecture and prevents the extension of parameters space which would be caused by extending to encoder-decoder structure in an auto-encoder setting.

We will show quantitatively that regression based pretraining improves prediction performance as well as classification performance for traversable terrain. Furthermore, we qualitatively present the advantages in the context of corner cases of manual cost functions, where pretraining followed by IRL-based training is able to recover more accurate and safe cost representations.

\section{Related Work}

A major share of early work in IRL focuses on small-scale scenarios and benchmarks \cite{ziebart2008maximum,ng2000algorithms}.
%However, as technology advances an increasing percentage of research in IRL starts targeting \dushyant{an odd expression}
However, with recent technological advances, IRL approaches have been applied to larger state and feature spaces in real life applications \cite{FinnLA16,WulfmeierIROS2016}.
In particular, these techniques harness the potential of deep neural networks, learning rich representations that are able to model the relationship between the state of the environment and the reward structure implied by demonstrated behaviour.

When working with large state spaces and end-to-end learning of reward functions, exploration of the state space becomes more important to enable learning rich feature representations and improve on generalisation.
Such capabilities are particularly important when learning driving behaviours from human demonstration: since the demonstrations cannot cover every possible driving scenario, it is necessary to ensure the vehicle can handle the unseen or scarcely seen ``corner cases''.

%In recent work~\cite{WulfmeierIROS2016}, we perform MaxEnt Deep IRL to learn policies from human demonstrations for autonomous driving applications.
%This approach exploits deep neural networks to learn rich representations that can accurately capture the reward structure.

% \markus{careful with such statements, because this is kind of what IRL inherently does}
% However, none of these techniques explore the capability to inject human domain knowledge into the learning process.

One paradigm that has been commonly employed in the past for deep neural networks is that of pretraining.
In previous work, networks are pretrained in a greedy layer-wise fashion~\cite{bengio2007greedy}, by training each layer as a single layer unsupervised model (such as an autoencoder or restricted Boltzmann machine~\cite{coates2010analysis}), and using the hidden activations as the input to the next layer of unsupervised training.
This pretraining has the effect of initialising the weights of the deep neural network to a region of the parameter space that is good for unsupervised tasks~\cite{erhan2010does}, from which the entire model can be fine-tuned to the specific classification task.
This can be seen as being conceptually similar to the notion of \emph{inductive transfer}, in which a model or representation learned for one task can be utilised or adapted to another.

In this work, we draw inspiration from these ideas, by performing pretraining of the deep IRL model with a manual cost map.
In contrast to the unsupervised layer-wise training approach, however, we pretrain the network as a \emph{regressor} to predict the manual cost map as output.
This has the effect of initialising the network weights to a region of the parameter space which can accurately represent the manual cost map.
This affords us the ability to inject domain knowledge into the network as a `human prior', which is a crucial capability for self-driving vehicles to ensure that environments and behaviours that are absent from the expert demonstration data are handled gracefully.

\section{Methods}
\subsection{Maximum Entropy Deep Inverse Reinforcement Learning} \label{sec:maxent}
The goal of IRL is to infer the reward structure that underlies certain behaviours.
%Here, we focus on the case in which the reward structure must be derived from expert demonstrations of a particular task.
The process can be defined under a Markov Decision Process framework $\mathcal{M} = \{\mathcal{S}, \mathcal{A}, \mathcal{T}, \gamma, r \}$, where $\mathcal{S}$ is the state space, $\mathcal{A}$ is the set of possible actions, $\mathcal{T}$ denotes the state transition model, $\gamma \in (0, 1]$ is a discount factor that moderates the influence of future rewards, and $r: \mathcal{S} \times \mathcal{A} \rightarrow \mathbb{R}$ is a function specifying the reward structure.
As $r$ is unknown, it must be inferred from a set of demonstrations $\mathcal{D} = \{\varsigma_1, \varsigma_2, \ldots, \varsigma_N\}$, each of which is a sequence of state-action pairs $\varsigma_i = \{ (s_1, a_1), (s_2, a_2), \ldots, (s_K, a_K)\}$ representing a sample trajectory.% driven by a human driver.

%The expert demonstration trajectories are obtained from large driving datasets which feature multiple human drivers with unspecified driving behaviour, which means that two key considerations need to be made.
%Firstly, not all of the demonstrations represent optimal behaviour, which means that the model of the expert needs to explicitly handle sub-optimality with respect to the reward.
%Secondly, there may be several optimal behaviours for a particular task, especially with multiple drivers, which needs to be reflected in the inferred reward structure.
The IRL model commonly needs to overcome two problems when reasoning about reward structures: suboptimality of sample trajectories given the underlying reward; and reward ambiguity, where multiple rewards can explain the same behaviour.

The Maximum Entropy (MaxEnt) approach to IRL explicitly addresses these concerns, by representing the expert behaviour as a distribution over demonstrated trajectories and assuming that this distribution has maximal entropy.
The agent's behaviour under the policy $\pi_{\mathcal{D}} (a \mid s)$ maximises the reward given the current model.
Consequently, the probability of any trajectory $\varsigma$ between specified initial and final states is proportional to the exponential of the reward along the path:
\begin{equation}
    P(\varsigma \mid r) = \prod_{i=1}^K \pi_D(a_i \mid s_i) \propto \exp \left\{ \sum_{{i=1}}^K  ~ r_{s_i,a_i} \right\}.
\label{eq:maxent_prob}
\end{equation}

The training loss in Maximum Entropy Deep Inverse Reinforcement Learning (MEDIRL) \cite{2015deepIRL} consists of a data term, which maximises the log likelihood of the demonstration trajectories given the parametrised reward function, and a regularisation term:
\begin{equation}
\mathcal{L(\theta)} = \log P(\mathcal{D},\theta \mid r(\theta)) = \underbrace{\log P(\mathcal{D} \mid r(\theta))}_{\mathcal{L_D}} + \underbrace{\log P(\theta)}_{\mathcal{L_\theta}}.
\label{eq:maxent_objective}
\end{equation}
% \noindent where the regulariser may be the $\ell_1$ or $\ell_2$ norms, or a combination of the two.

% The fully differentiable nature of the loss function lends itself to training neural network architectures, as the gradients could be backpropagated through the layers of the network.
%We exploit a convolutional neural network (CNN) to predict $r(\theta)$ from the raw laser input, enabling the ability to learn rich representations that capture the complex relationship between obstacles in the environment and the reward structure implied by human demonstrations.

The data term in Equation~\ref{eq:maxent_objective} can be split using the chain rule, into the gradient of the objective with respect to the reward, and the gradient of the reward with respect to the network parameters:
\begin{eqnarray}\label{eq:maxent_gradients}
\frac{\partial \mathcal{L_D}}{\partial \theta} &=& \frac{ \partial \mathcal{L_D} }{ \partial r } \frac{ \partial r }{ \partial \theta }\\ \nonumber
&=& \underbrace{\left(\mu_{\mathcal{D}} - \mathbb{E} [\mu]\right)}_{ {State Visitation} \atop {Frequency Matching}} \underbrace{\frac{\partial}{\partial \theta} r(\theta)}_{ {Backpropagation}} .
\end{eqnarray}
\noindent Here, the first term computes the difference in state visitation frequencies - how often specific states are traversed - between the demonstrations ($\mu_{\mathcal{D}}$) and the model; while the second term backpropagates this error through the network. It becomes obvious in this formulation that the area and therefore the types of features learned based on these error terms focus around the demonstration trajectories.

\subsection{Model Architecture}
The CNN architecture used for this work is the multi-scale fully convolutional network (MS-FCN) proposed in~\cite{WulfmeierIROS2016}.
The input to the network is a 2D laser occupancy grid, and the output is a cost map specifying the reward for each location in the grid.
The architecture consists of $5\times5$, $3\times3$, and $1\times1$ convolutional layers, including nonlinearities, max pooling and upsampling layers, and a separate branch to learn intermediate representations at a different scale, which are concatenated in the higher layers.
For more details, the reader is referred to~\cite{WulfmeierIROS2016}.

\subsection{Training}
Each iteration of training under the MEDIRL framework is composed of the following steps.
First, the reward function (cost map) is computed by performing a forward pass of the network, and the MDP is solved for this current reward estimate.
Using the policy $\pi$ computed by the MDP, we can obtain the expected state visitation frequencies $\mathbb{E} [\mu]$ and the MaxEnt loss and gradients from Equations \ref{eq:maxent_objective} and \ref{eq:maxent_gradients}.
Finally, these errors are backpropagated through the network to obtain the weight updates.

In the pretraining setup, the CNN first learns to predict the manual cost map as output - which results in a dense map of gradients over the full area - and then fine-tuned with the above steps.
This is discussed in further detail in the next section.

\subsection{Incorporating Human Priors}

\begin{figure}[h]
	\centering
	\includegraphics[width = \textwidth]{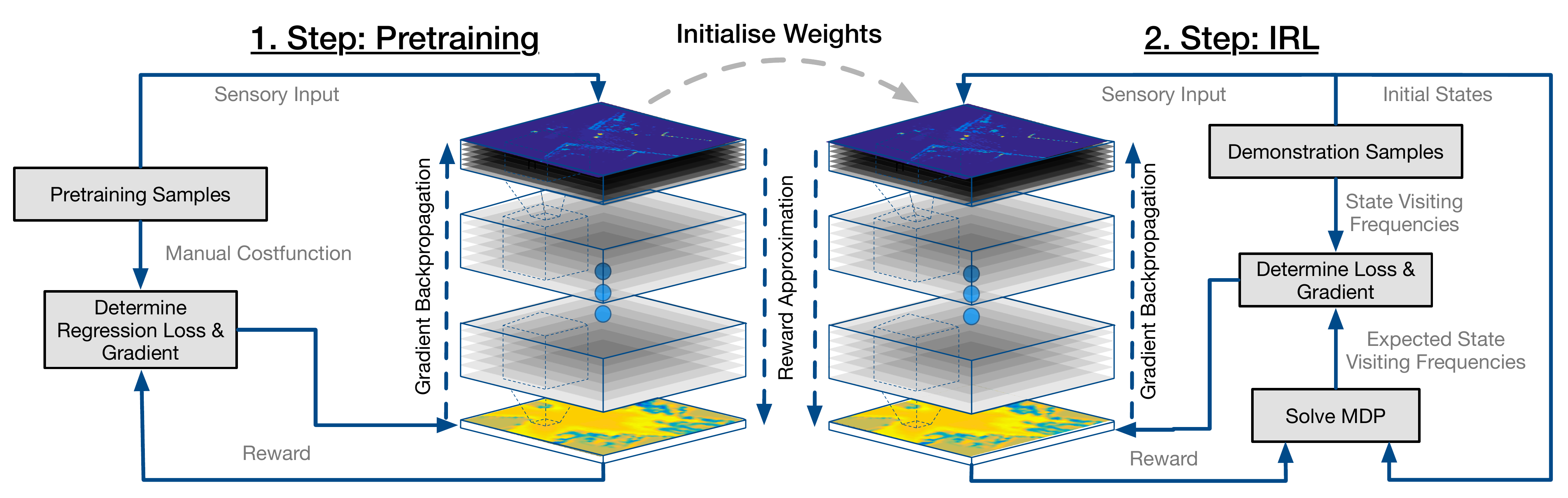}
 	\caption{ Schema for additional network pretraining, where the model learns to regress to a manual prior cost map. Subsequently the network is fine-tuned to predict the reward under the MEDIRL framework.}
	\label{fig:pretrain}
\end{figure}

Due to the iterative nature of planning and refining the cost model in the training process, the approach described in Section \ref{sec:maxent} tends to focus on discriminative features around highly visited states as depicted in Figure \ref{fig:sparse}. Features for terrains that are neither explored by demonstration samples nor the planning step will never be formed by backpropagation of error terms through the network.
Therefore, the model has to generalise based on similarity to more commonly traverse areas, resulting in artifacts and noisy reward maps. This situation is enhanced in scenarios where feature representations for similar places differ based on their position in state space, as is the case in the dataset used in this work \cite{WulfmeierIROS2016}. The LIDAR scan points in this setup will be spread sparser at greater distance from the car, resulting in spatially variant representation.

In order to address this shortcoming, we suggest training the network as a regressor towards a prior cost map. 
%While we train the network for the same task of producing a cost map, the training procedure differs significantly - framing the full two step method in the context of transfer learning.
These cost maps can be automatically generated from the laser input data based on manually handcrafted features, which enables us to utilise the availability of large amounts of data without human labelling efforts.
However, the principal benefit is the ability to explore all features relevant to generating the cost function, leading to better generalisation in areas with greater distance from the demonstration trajectories.

\section{Experiments}
We evaluate all approaches on the large scale dataset presented in \cite{WulfmeierIROS2016} and investigate how performance progresses between training from random initialisation and employing prior human knowledge. 

\subsection{Data}
The dataset consists of 25,000 demonstration samples of urban driving from 13 different drivers. The data was collected in the inner city of Milton Keynes and includes driving around different types of obstacles, including but not restricted to: bollards, green patches, bike racks, slopes, cars, and underpasses.
To increase the performance of our approaches as path planning cost maps, we added common preprocessing steps, such as normalising the input data, and trained on shorter trajectories that are more representative for the motion primitives employed in our motion planning framework.

% In addition, we increase the number and diversity - in terms of location and shape - of positive and negative example trajectories for the classification task in Section \ref{sec:classif} to gain a more robust metric for evaluating their performance.

\subsection{Prediction Performance}
To evaluate how well the trained models approximate human behaviour, we use two common metrics: the negative log-likelihood (NLL) and the modified Hausdorff Distance (MHD) \cite{kitani_activity_2012}. The first metric is representing how likely the expert demonstrations are given the current cost function and the latter is a spatial metric for how close the demonstrations are to samples drawn from a probability distribution given the cost map. The learned models generally outperform the handcrafted approach as increasing the probability of the demonstration samples is inherently part of their objective function, in contrast to the manual cost function. As Table \ref{tab:metrics} displays pretraining improves our ability to predict where people are more likely to drive.

\begin{table}[!h]
    \centering
    \begin{tabular}{l|M{30mm}|M{30mm}}
       \toprule
        Metric & NLL & MHD \\
        \midrule
        Manual cost function & 56.402 & 0.286  \\
        wo pretraining & 47.535 & 0.218  \\
        w pretraining & \textbf{46.767} & \textbf{0.182}  \\
        \bottomrule
    \end{tabular}
    \vspace{3pt}
    \caption{Evaluation of cost functions for urban driving under the negative log-likelihood (NLL) and Modified Hausdorff Distance (MHD) metrics. Lower numbers represent models that are approximating human behaviour with higher precision.}
    \label{tab:metrics}
\end{table}

\subsection{Classification Performance}\label{sec:classif}
One major drawback for evaluating all approaches in a real world setup is the absence of absolute ground truth for the cost map. To overcome this impediment, we evaluate the approach based on its performance as a classifier for feasible example trajectories. Given a set of traversable trajectories - taken from our test set - and a set of artificial collision trajectories, we can analyse the accuracy of the model in a classification setup for traversable trajectories.
In this setup, traversable terrain is taken as the positive class, meaning that a high precision model is conservative when assigning traversability.

\begin{figure}[h]
	\centering
	\includegraphics[width = 0.4\textwidth]{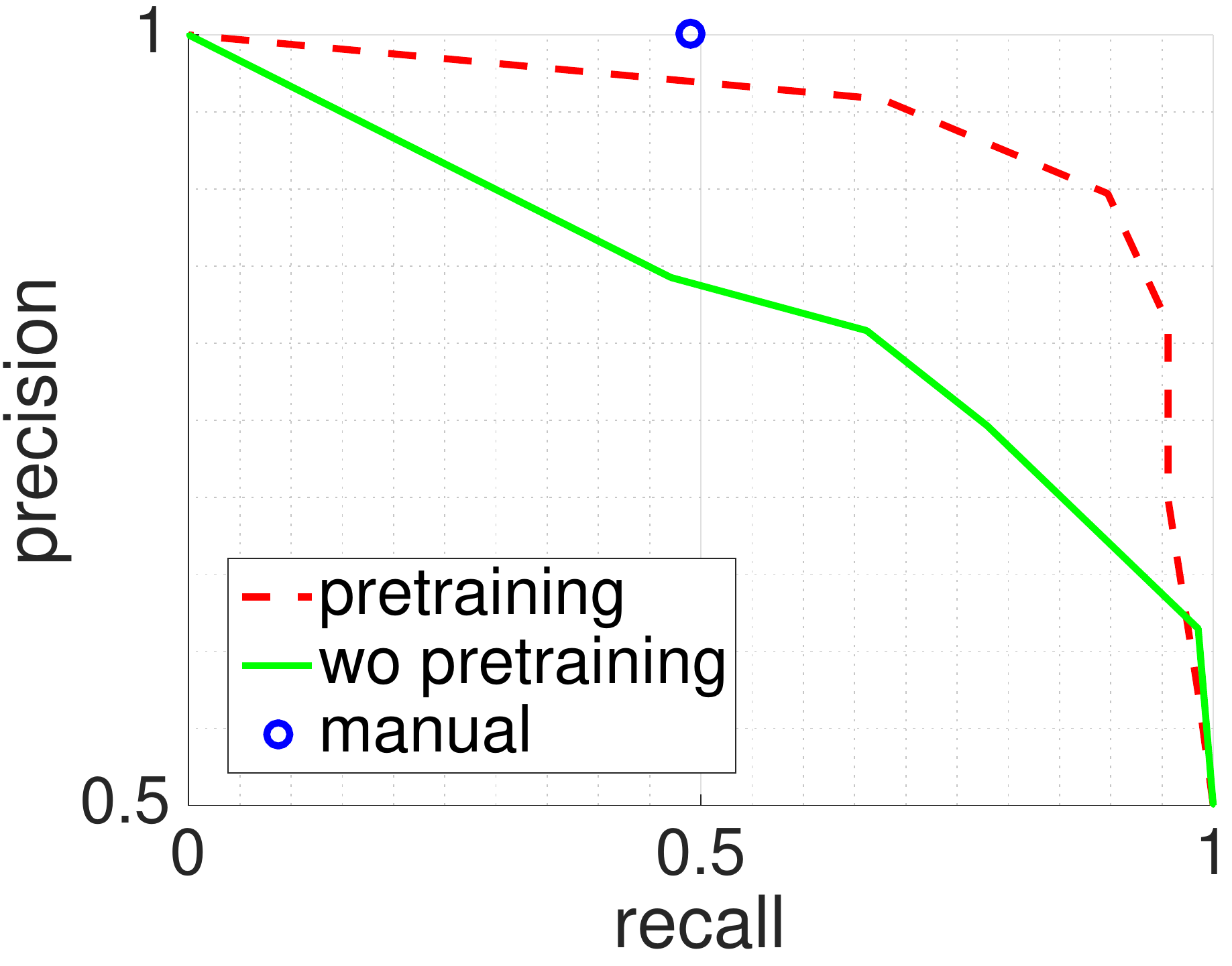}
	\caption{Precision Recall (PR) Curves for Trajectory Classification. The manual cost function has high precision but low recall, meaning that it is safe but conservative and will falsely classify a significant number of feasible trajectories as untraversable. Applying human priors in the pretraining step enables a significant gain in precision towards the baseline. This method approaches the precision of the manual cost function while strongly exceeding it in recall.}
	\label{fig:precisrecall}
\end{figure}

While utilising human priors already improves accuracy for prediction, its principal gain lies in being able to improve spatial generalisation and the ability to learn more robust cost functions. %, as displayed in Table \ref{tab:cornercases}.
Figure \ref{fig:precisrecall} depicts the Precision-Recall curves for networks with and without pretraining. The handcrafted cost function does not include a threshold parameter and is therefore represented as a point. While this approach is manually designed to be conservative and enables us to operate at maximum precision, it falsely rejects much of the terrain as untraversable. The learned cost functions enable us to find possible paths in many situations when the manual cost function will get stuck. When introducing human prior knowledge into the training process, the approach achieves a significant gain in precision compared to random initialisation. Hence, utilising this knowledge is an important step towards robust application of learned cost maps. 

\begin{table}[H]

    \centering
    \begin{tabular}{lM{39mm}M{39mm}M{39mm}}
      \toprule
        Scenario & Manual cost function & wo pretraining & w pretraining \\
        \midrule
        Stairs & 
        \includegraphics[width=\linewidth]{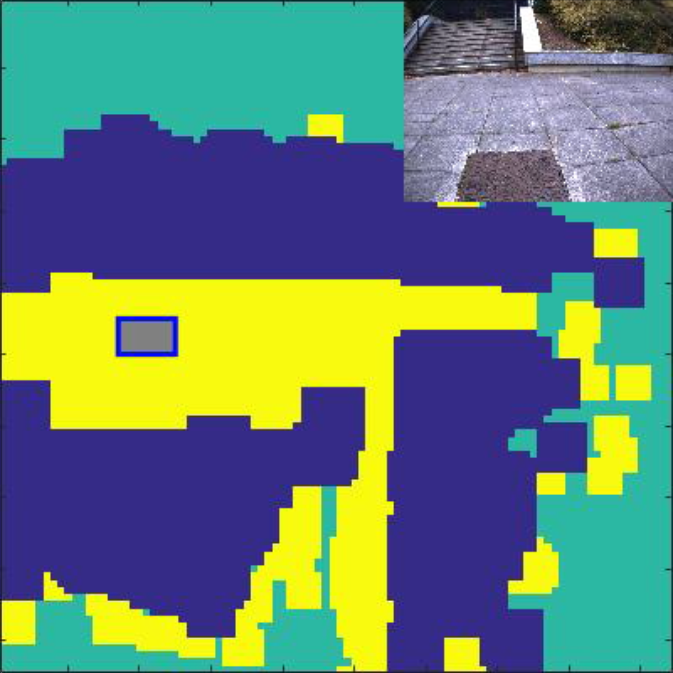} & 
        \includegraphics[width=\linewidth]{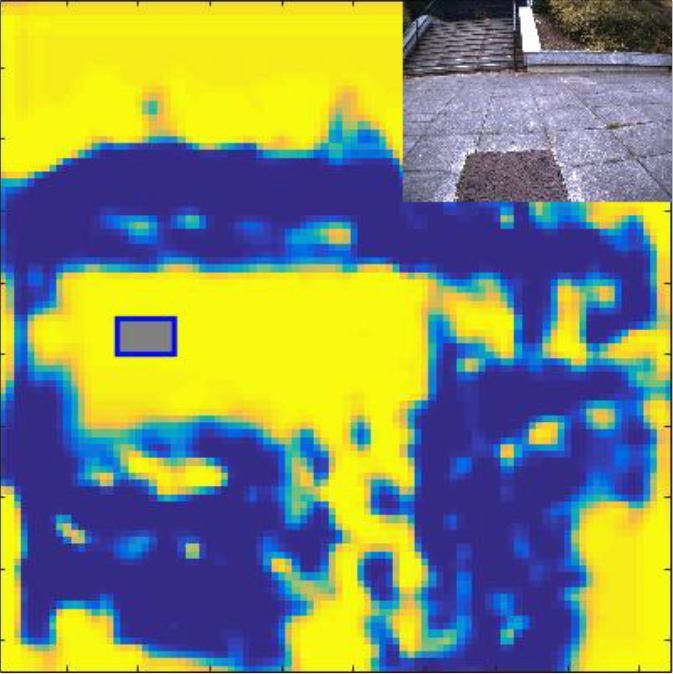} & 
        \includegraphics[width=\linewidth]{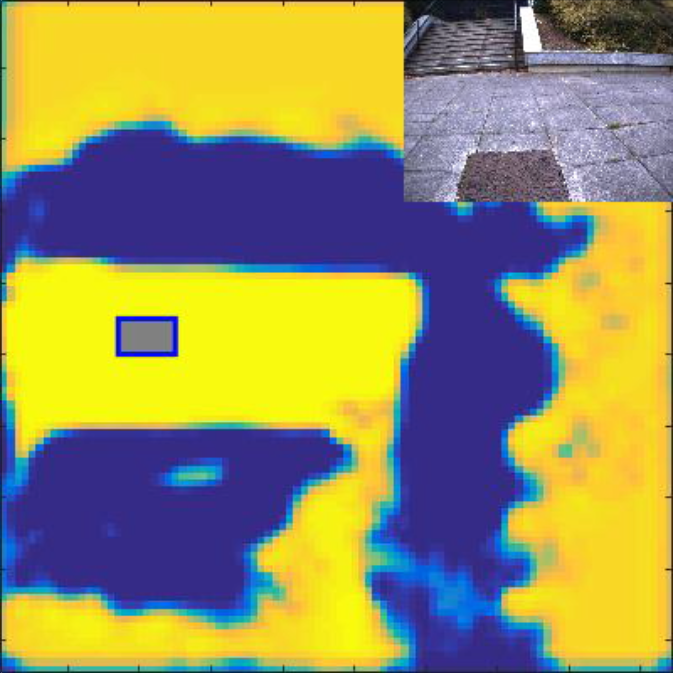} \\  
        Bollards &
        \includegraphics[width=\linewidth]{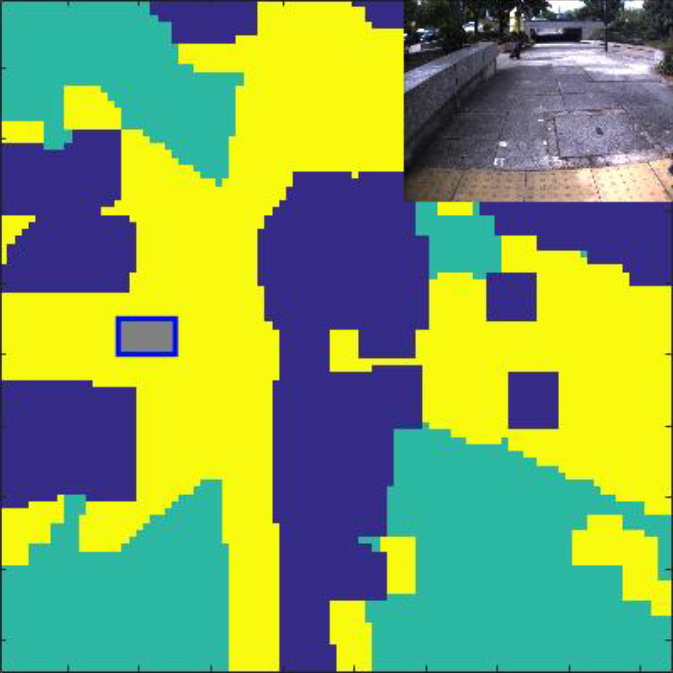} & 
        \includegraphics[width=\linewidth]{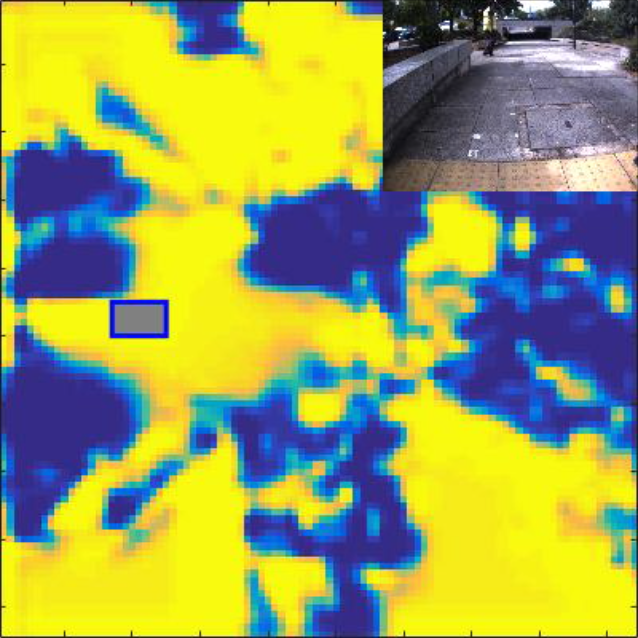} & 
        \includegraphics[width=\linewidth]{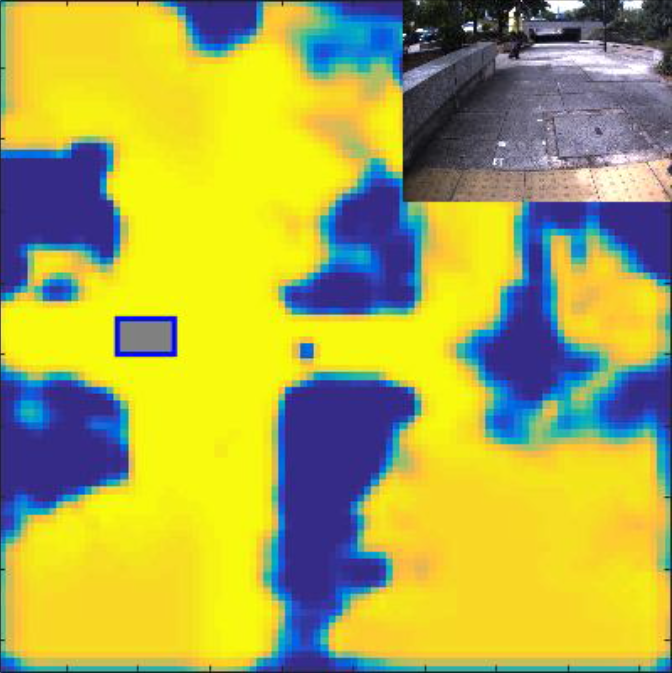} \\
        Grass &
        \includegraphics[width=\linewidth]{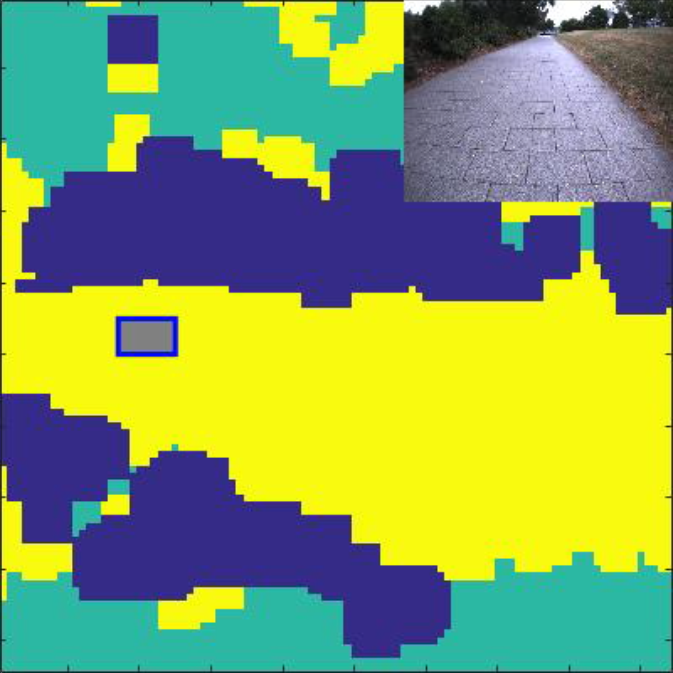} & 
        \includegraphics[width=\linewidth]{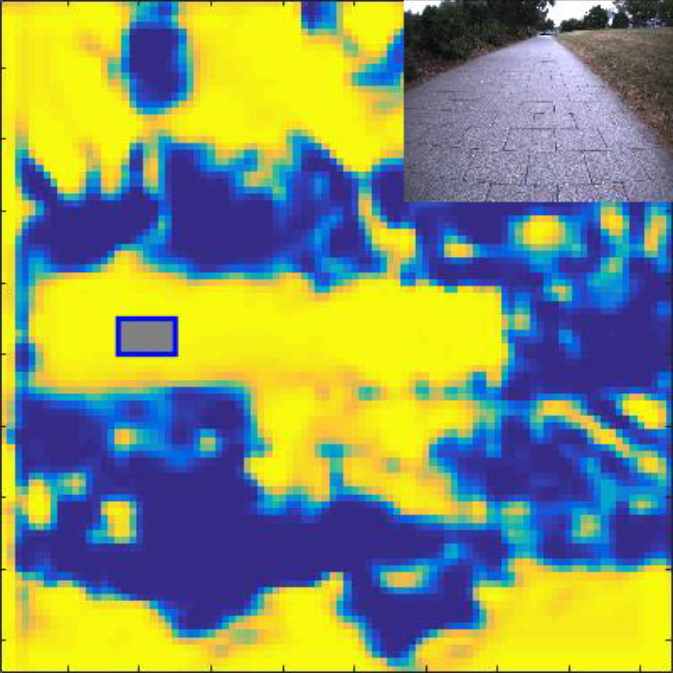} & 
        \includegraphics[width=\linewidth]{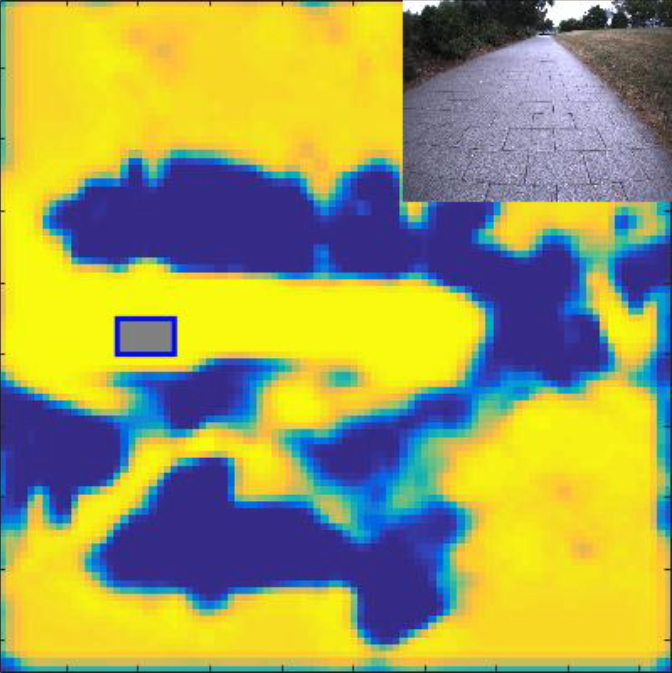} \\
        Underpass & 
        \includegraphics[width=\linewidth]{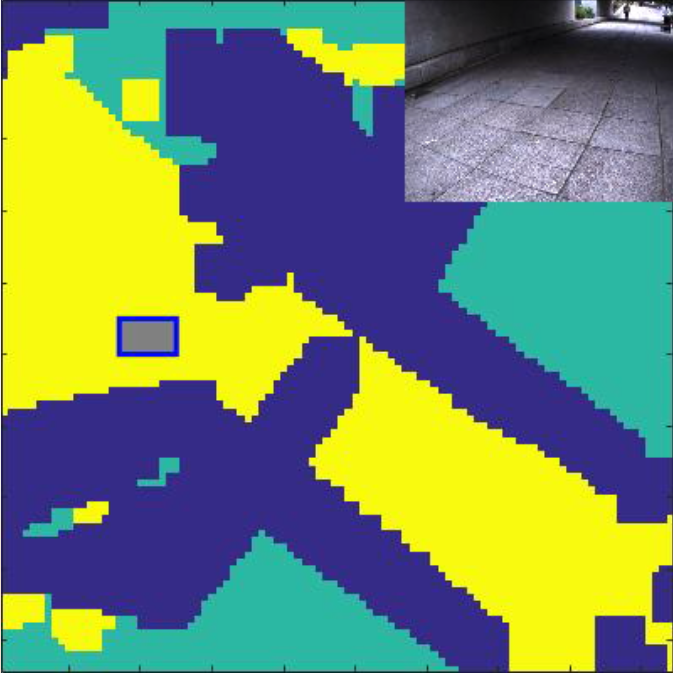} & 
        \includegraphics[width=\linewidth]{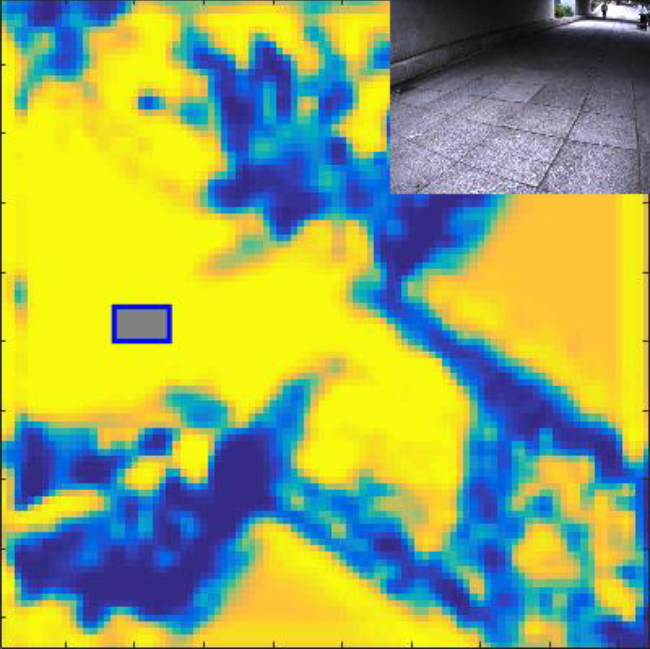} & 
        \includegraphics[width=\linewidth]{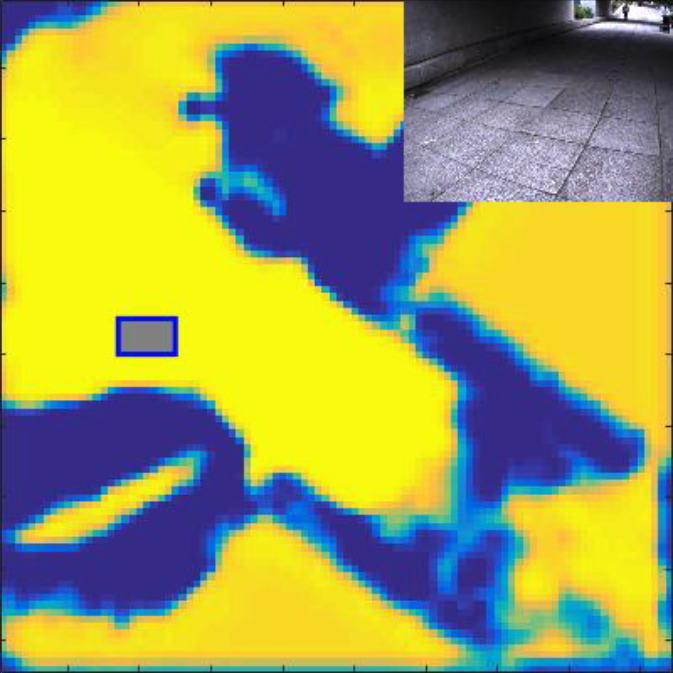} \\
        Slope & 
        \includegraphics[width=\linewidth]{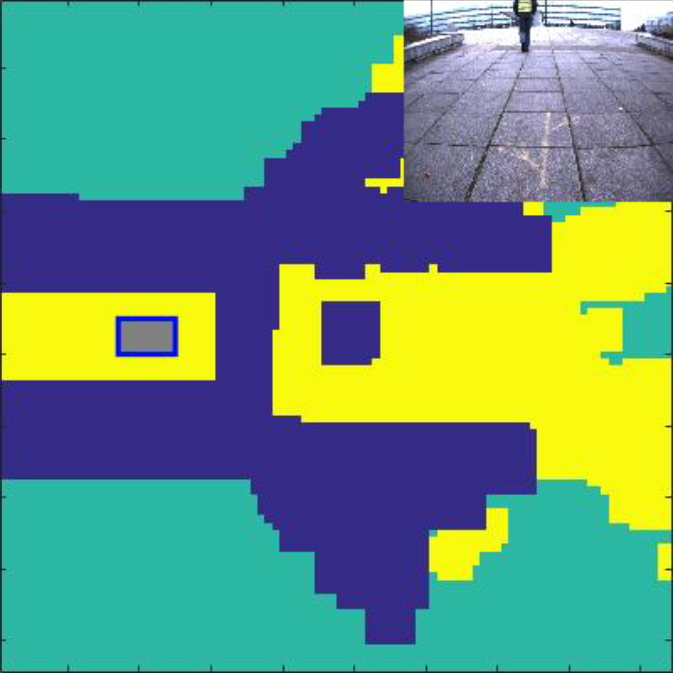} & 
        \includegraphics[width=\linewidth]{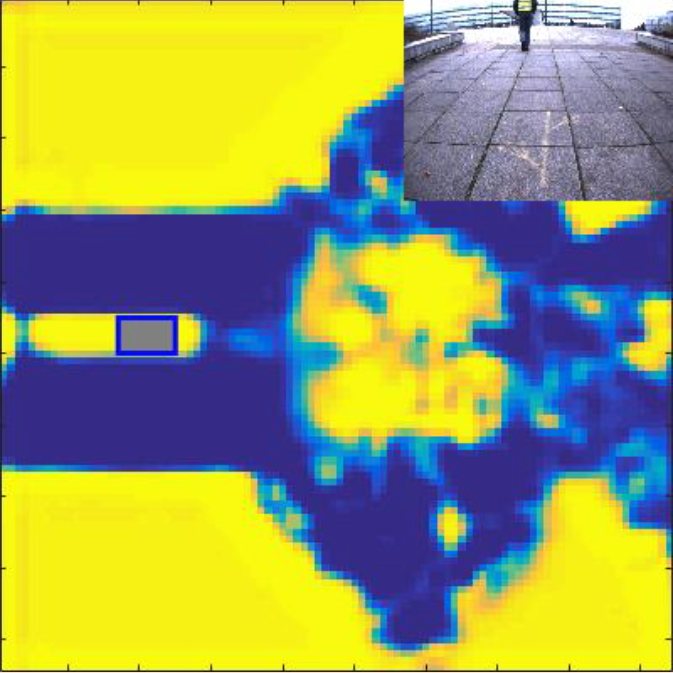} & 
        \includegraphics[width=\linewidth]{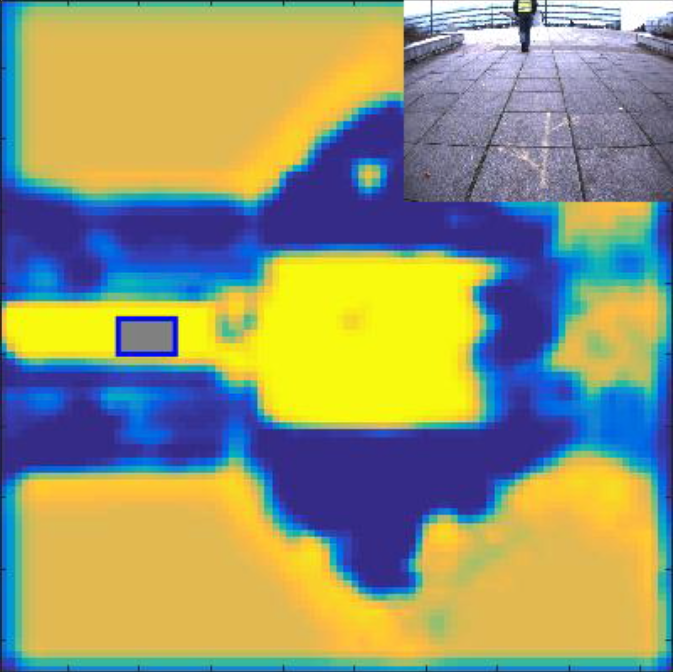} \\
        \bottomrule
    \end{tabular}
    \vspace{3pt}
    \caption{Corner Cases for the Cost Function. The photos include views from the front facing camera module with the vehicle represented gray rectangle driving towards the right side of each cost map. Obstacles are represented in blue while yellow depicts traversable terrain.}
    \label{tab:cornercases}
\end{table}

\subsection{Qualitative Assessment of Learned Cost Functions}

Table \ref{tab:cornercases} represents various situations emphasising shortcomings of the manual cost function. 
One point of high importance is, that all learned cost functions show additional obstacles starting at about 13 meters distance from the vehicle position, which is the length of the demonstration trajectories. By choosing the length of the trajectories, we define how far we trust our perception systems. Since features in distant areas are only traversed but planner and not demonstration samples in the training process, they will classified as untraversable with high probability.

The principal rules behind the handcrafted cost function are based on a threshold on the height range of detected points within a cell and the expansion of obstacles by the size of the vehicle to enable point based planning. This can lead to inaccuracies in the presence of slopes, which can exceed the threshold and will be shown as untraversable, and the same can occur for underpasses, where scans from ceiling and floor result in a high height range. Stairs on the other hand can still fit within the same threshold but present obstacles for any vehicle since they cannot be traversed due to their indiscontinuity. Bollards that are extended slightly too far will seem untraversable and areas such as grass might look very similar in features to pathways but should not be traversed. 

While the randomly initialised network already learns to represent the main obstacles and traversable areas, it results in some noisy areas and artificial obstacles. Based on human prior domain knowledge, the network learns to refine the representation and is significantly more robust, learns to represent distinct obstacle boundaries and displays fewer artifacts. The approach learns that slopes are traversable, while stairs are not and extends obstacle boundaries only as far as necessary for safe traversal as seen in the respective cases in Table \ref{tab:cornercases}.

We conjecture that without pretraining, a lot of the expressive power of the network is employed to learn the very specific representation focused around the demonstration trajectories. If we instead pretrain the model to predict an existing cost map, the remainder of the learning process is able to generalise broader and better capture some of the corner cases described previously.

\subsection{Implementation Details}

When performing path planning, a threshold has to be determined to define untraversable and unsafe terrain. To simplify this step we normalise the output by sending it through a sigmoid. After first trials with Rectified Linear Units (ReLU) as activation functions within the network, the sigmoid saturates early on and training slows down, we exchanged all activation functions with sigmoid units.
Furthermore, we introduce early stopping to increase generalisation performance; since following the pretraining step, the model requires fewer iterations to converge.
%Furthermore, we introduce early stopping to increase generalisation performance. Early stopping gains particularly importance in this context as - following our pretraining - we need fewer iterations to converge.

\section{Conclusions and Ongoing Work}
We develop an approach for incorporating human prior knowledge into cost function learning in the context of IRL.
By utilising human priors, we can improve on prediction of demonstrator trajectories. Adding the pretraining step enables us to start from a significantly more desirable network initialisation; giving a richer representation for features at arbitrary position and enabling more effective generalisation. The approach improved classification performance for traversable terrain and results in more robust and distinct cost representations. Our evaluation furthermore depicts the advantages in specific corner cases of our environment given by slopes, bollards, underpasses and stairs. 

% \begin{figure}[h]
% 	\centering
% 	\includegraphics[width = 0.4\textwidth]{Figures/pronet.pdf}
% 	\caption{Architecture for a progressive net-like approach that could exploit human priors in IRL.}
% 	\label{fig:pronet}
% \end{figure}

%\markus{I would prefer scientific discourse at the workshop about it. plus there is a high probability that we wont be able to follow up on it due to other projects}\dushyant{ok, but you still need to beef up the first paragraph, there's nothing about how the technique works or what the results indicate. "improve prediction and classification performance" is too vague. be more specific.} Ok beefing up now
Ongoing research looks into employing variations of Progressive Neural Networks \cite{prognets16}%as displayed in Figure %\ref{fig:pronet} 
, which represent another opportunity to exploit human priors by switching to modelling the error/residual between the human given prior and the true cost function. 
Different variations of the approach can focus on directly reusing the learned features of the trained model or simply focus on approximating the difference between the manual cost function and the optimal function to describe human behaviour.

The focus shifts in context of the latter from learning all representations needed to approximate a function to learning the difference between the manual cost function and the one best suited to explain human behaviour. This approach benefits from expert domain knowledge such that it only needs to embody the variations that the expert did not consider. 
%However, if the resulting difference function presents patterns of higher complexity and more abrupt variations, this procedure can more difficult. 

\section{Acknowledgements}
The authors would like to acknowledge the support of this work by the EPSRC through grant number DFR01420, and a DTA Studentship.

% \section*{References}

\bibliographystyle{unsrt}
\bibliography{main.bib}

\end{document}